\documentclass{llncs}
\pdfoutput=1
\usepackage{amssymb}
\usepackage{makeidx}  
\usepackage{url}     
\usepackage{amsmath}
\usepackage{nccmath}
\usepackage{amsfonts} 
\usepackage{graphicx}
\usepackage{multirow}
\usepackage{bm}

\usepackage{array,booktabs,arydshln,xcolor}
\usepackage{cite}
\usepackage{floatrow}
\usepackage{pbox}
\newcolumntype{M}[1]{>{\centering\arraybackslash}m{#1}}

\begin{document}
\frontmatter          
\pagestyle{headings}  
\mainmatter              
\title{Generalised Dice overlap as a deep learning loss function for highly unbalanced segmentations}%
\author{Carole H. Sudre\inst{1}${}^{,}$\inst{2} \and Wenqi Li\inst{1} \and Tom Vercauteren\inst{1} \and Sebastien Ourselin\inst{1}${}^{,}$\inst{2}  \and M. Jorge Cardoso \inst{1}${}^{,}$\inst{2}}
\institute{Translational Imaging Group, CMIC, University College London, NW1 2HE, UK \and Dementia Research Centre, UCL Institute of Neurology, London, WC1N 3BG, UK } 
 
\maketitle
\begin{abstract}
Deep-learning has proved in recent years to be a powerful tool for image analysis and is now widely used to segment both 2D and 3D medical images. Deep-learning segmentation frameworks rely not only on the choice of network architecture but also on the choice of  loss function. When the segmentation process targets rare observations, a severe class imbalance is likely to occur between candidate labels, thus resulting in sub-optimal performance. In order to mitigate this issue, strategies such as the weighted cross-entropy function, the sensitivity function or the Dice loss function, have been proposed. In this work, we investigate the behavior of these loss functions and their sensitivity to learning rate tuning in the presence of different rates of label imbalance across 2D and 3D segmentation tasks. We also propose to use the class re-balancing properties of the Generalized Dice overlap, a known metric for segmentation assessment, as a robust and accurate deep-learning loss function for unbalanced tasks.  
\end{abstract}
\section[Introduction]{Introduction}
\label{sec:intro}
\sectionmark{Introduction}
A common task in the analysis of medical images is the ability to detect, segment and characterize pathological regions that represent a very small fraction of the full image. This is the case for instance with brain tumors or white matter lesions in multiple sclerosis or aging populations. 
Such unbalanced problems are known to cause instability in well established, generative and discriminative, segmentation frameworks. 
Deep learning frameworks have been successfully applied to the segmentation of 2D biological data and more recently been extended to 3D problems \cite{Zheng2015}. Recent years have seen the design of multiple strategies to deal with class imbalance (e.g. specific organ, pathology...). Among these strategies, some focus their efforts in reducing the imbalance by the selection of the training samples being analyzed at the risk of reducing the variability in training  \cite{Havaei,Lai2015}, while others have derived more appropriate and robust loss functions \cite{Ronneberger2015,Brosch2015,Milletari2016}.
In this work, we investigate the training behavior of three previously published loss functions in different multi-class segmentation problems in 2D and 3D while assessing their robustness to learning rate and sample rates. We also propose to use the class re-balancing properties of the Generalized Dice overlap as a novel loss function for both balanced and unbalanced data.  

\section[Methods]{Methods}
\sectionmark{Methods}
\subsection{Loss functions for unbalanced data}
The loss functions compared in this work have been selected due to their potential to tackle class imbalance. All loss functions have been analyzed under a binary classification  (foreground vs. background) formulation as it represents the simplest setup that allows for the quantification of class imbalance. Note that formulating some of these loss functions as a 1-class problem would mitigate to some extent the imbalance problem, but the results would not generalize easily to more than one class. Let $R$ be the reference foreground segmentation (gold standard) with voxel values $r_n$, and $P$ the predicted probabilistic map for the foreground label over $N$ image elements $p_n$, with the background class probability being $1-P$. 
\paragraph{\textbf{Weighted cross-entropy (WCE):}} The weighted cross-entropy has been notably used in \cite{Ronneberger2015}. The two-class form of WCE can be expressed as \[\text{WCE}=-\dfrac{1}{N}\sum_{n=1}^{N}wr_n\log(p_n)+(1-r_n)\log(1-p_n),\]\noindent where $w$ is the weight attributed to the foreground class, here defined as $w=\frac{N-\sum_{n}p_n}{\sum_{n}p_n}$. The weighted cross-entropy can be trivially extended to more than two classes.
\paragraph{\textbf{Dice loss (DL)}} The Dice score coefficient (DSC) is a measure of overlap widely used to assess segmentation performance when a gold standard or ground truth is available. Proposed in Milletari et al. \cite{Milletari2016} as a loss function, the 2-class variant of the Dice loss, denoted $DL_2$, can be expressed as \[\text{DL}_2=1-\dfrac{\sum_{n=1}^{N}p_{n}r_{n}+\epsilon}{\sum_{n=1}^{N}p_n+r_n+\epsilon}-\dfrac{\sum_{n=1}^{N}(1-p_{n})(1-r_{n})+\epsilon}{\sum_{n=1}^{N}2-p_n-r_n+\epsilon}\]
The $\epsilon$ term is used here to ensure the loss function stability by avoiding the numerical issue of dividing by 0, i.e. $R$ and $P$ empty.
\paragraph{\textbf{Sensitivity - Specificity (SS):}} Sensitivity and specificity are two highly regarded characteristics when assessing segmentation results. The transformation of these assessments into a loss function has been described by Brosch et al. \cite{Brosch2015} as \[\text{SS}=\lambda\dfrac{\sum_{n=1}^{N}(r_n-p_n)^2r_n}{\sum_{n=1}^{N}r_n+\epsilon}+(1-\lambda)\dfrac{\sum_{n=1}^{N}(r_n-p_n)^2(1-r_n)}{\sum_{n=1}^{N}(1-r_n)+\epsilon}.\] The parameter $\lambda$, that weights the balance between sensitivity and specificity, was set to 0.05 as suggested in \cite{Brosch2015}. The $\epsilon$ term is again needed to deal with cases of division by 0 when one of the sets is empty. 
\paragraph{\textbf{Generalized Dice Loss (GDL):}}
Crum et al \cite{Crum2006} proposed the Generalized Dice Score (GDS) as a way of evaluating multiple class segmentation with a single score but has not yet been used in the context of discriminative model training. We propose to use the GDL as a loss function for training deep convolutional neural networks. It takes the form:
\[ \text{GDL}=1-2\dfrac{\sum_{l=1}^{2}w_l\sum_{n}r_{ln}p_{ln}}{\sum_{l=1}^{2}w_l\sum_{n}r_{ln}+p_{ln}}, \]
where $w_l$ is used to provide invariance to different label set properties. In the following, we adopt the notation $\text{GDL}_v$ when $w_l=1/(\sum_{n=1}^{N}r_{ln})^2$. As stated in \cite{Crum2006}, when choosing the $\text{GDL}_v$ weighting, the contribution of each label is corrected by the inverse of its volume, thus reducing the well known correlation between region size and Dice score. In terms of training with stochastic gradient descent, in the two-class problem, the gradient with respect to $p_{i}$ is:
\[\dfrac{\partial \text{GDL}}{\partial p_{i}}=-2\dfrac{(w_1^2-w_2^2)\left[\displaystyle\sum_{n=1}^{N}p_{n}r_{n}-r_{i}\sum_{n=1}^{N}(p_{n}+r_{n})\right]+Nw_2(w_1+w_2)(1-2r_{i})}{\left[(w_1-w_2)\sum_{n=1}^{N}(p_{n}+r_{n})+2Nw_2\right]^2}\]
Note that this gradient can be trivially extended to more than two classes. 
\subsection{Deep learning framework}
To extensively investigate the loss functions in different network architectures, four previously published networks were chosen as representative networks for segmentation due to their state-of-the art performance and were reimplemented using Tensorflow. 
\paragraph{\textbf{2D networks:}} Two networks designed for 2D images were used to assess the behaviour of the loss functions: UNet \cite{Ronneberger2015}, and the TwoPathCNN \cite{Havaei}.  The UNet architecture presents a U-shaped pattern where a step down is a series of two convolutions followed by a downsampling layer and a step up consists in a series of two convolution followed by upsampling. Connections are made between the downsample and upsample path at each scale. TwoPathCNN \cite{Havaei}, designed for tumor segmentation, is used here in a fully convolutional 2D setup under the common assumption that a 3D segmentation problem can be approximated by a 2D network in situations where the slice thickness is large. This network  involves the parallel training of two networks - a local and a global subnetwork. The former consists of two convolutional layers with kernel of size $7^2$ and $5^2$ with max-out regularization interleaved with max-pooling layers of size $4^2$ and $2^2$ respectively; while the latter network consists of a convolution layer of kernel size $13^2$ followed by a max-pooling of size $2^2$.  The features of the local and global networks are then concatenated before a final fully connected layer resulting in the classification of the central location of the input image. 
\paragraph{\textbf{3D networks:}} 
The DeepMedic architecture \cite{Kamnitsas2017} and the HighResNet network \cite{Li2017} were used in the 3D context. DeepMedic consists in the parallel training of one network considering the image at full resolution and another  on the downsampled version of the image. The resulting features are concatenated before the application of two fully connected layers resulting in the final segmentation.
HighResNet is a compact end-to-end network mapping an image volume to a voxel-wise segmentation with a successive set of convolutional blocks and residual connections.
To incorporate image features at multiple scales, the convolutional kernels are dilated with a factor of two or four. The spatial resolution of the input volume is maintained throughout the network.  

\section[Experiments and Results]{Experiments and Results}
\label{sec:exp}
\sectionmark{Experiments and Results}
\subsection{Experiments}
The two segmentation tasks we choose to highlight the impact of the loss function target brain pathology: the first task tackles tumor segmentation, a task where tumor location is often unknown and size varies widely, and the second comprises the segmentation of age-related white matter hyperintensities, a task where the lesions can present a variety of shapes, location and size. 

In order to assess each loss function training behavior, different sample and learning rates were tested for the two networks. The learning rates (LR) were chosen to be log-spaced and set to $10^{-3}$, $10^{-4}$ and $10^{-5}$. For each of the networks, three patch sizes (small:S, moderate:M, large:L), resulting in different effective field of views according to the design of the networks were used to train the models. A different batch size was used according to the patch size. Initial and effective patch sizes, batch size  and resulting imbalance for each network are gathered in Table \ref{tab:SamplingRate}.
 \begin{table}[pb]
 	\includegraphics[width=0.95\textwidth]{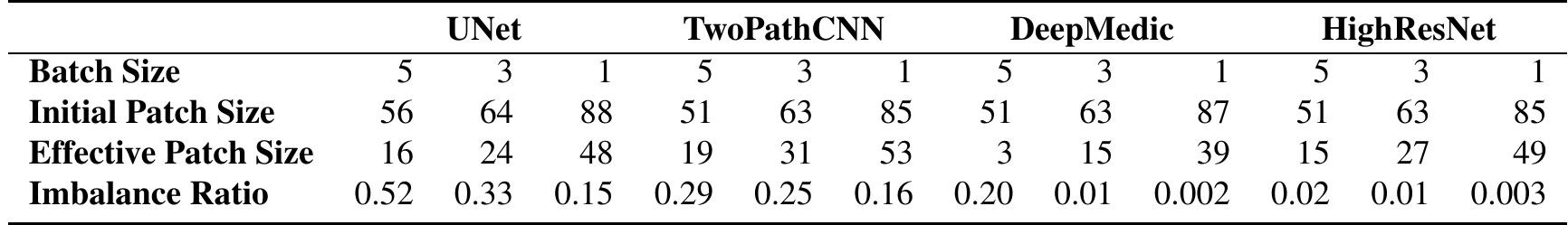}
 	\caption{Comparison of patch sizes and sample rate for the four networks.}
 	\label{tab:SamplingRate}
 \end{table}
 In order to ensure a reasonable behavior of all loss functions, training patches were selected if they contained at least one foreground element. Larger patch sizes represent generally more unbalanced training sets. The networks were trained without training data augmentation to ensure more comparability between training behaviors. The imbalance in patches varied greatly according to networks and contexts reaching at worst a median of 0.2\% of a 3D patch 
%

The 2D networks were applied to BRATS \cite{Menze2015}, a neuro-oncological dataset 
where the segmentation task was here to localize the background (healthy tissue) and the foreground (pathological tissue, here the tumor) in the image. 
The 3D networks were applied to an in house dataset of 524 subjects presenting age-related white matter hyperintensities. In both cases, the T1-weighted, T2-weighted and FLAIR data was intensity normalized by z-scoring the data according to the WM intensity distribution. 
The training was arbitrarily stopped after 1000 (resp. 3000) iterations for the 2D (resp. 3D) experiments, as it was found sufficient to allow for convergence for all metrics. 

 \subsection{2D Results}
   \begin{table}[pt]
   	\includegraphics[width=0.95\textwidth]{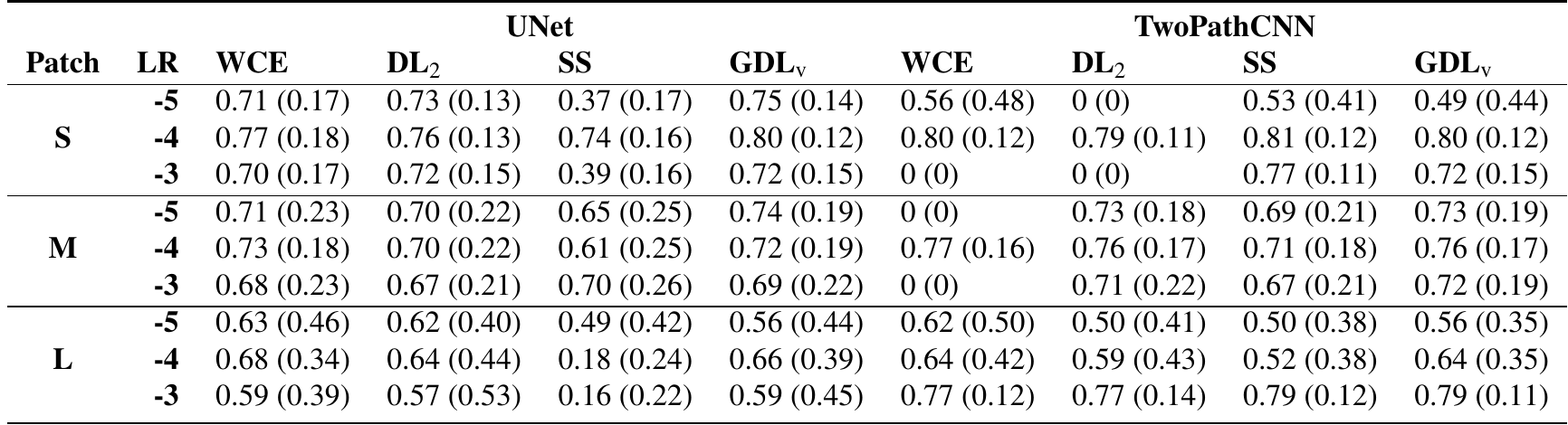}
   	\caption{Comparison of DSC over 200 last iterations in the 2D context for UNet and TwoPathCNN. Results are under the format median (interquartile range).}
   	\label{tab:2DResults}
   \end{table}
   \begin{figure}[pb!]
   	\includegraphics[width=0.9\textwidth]{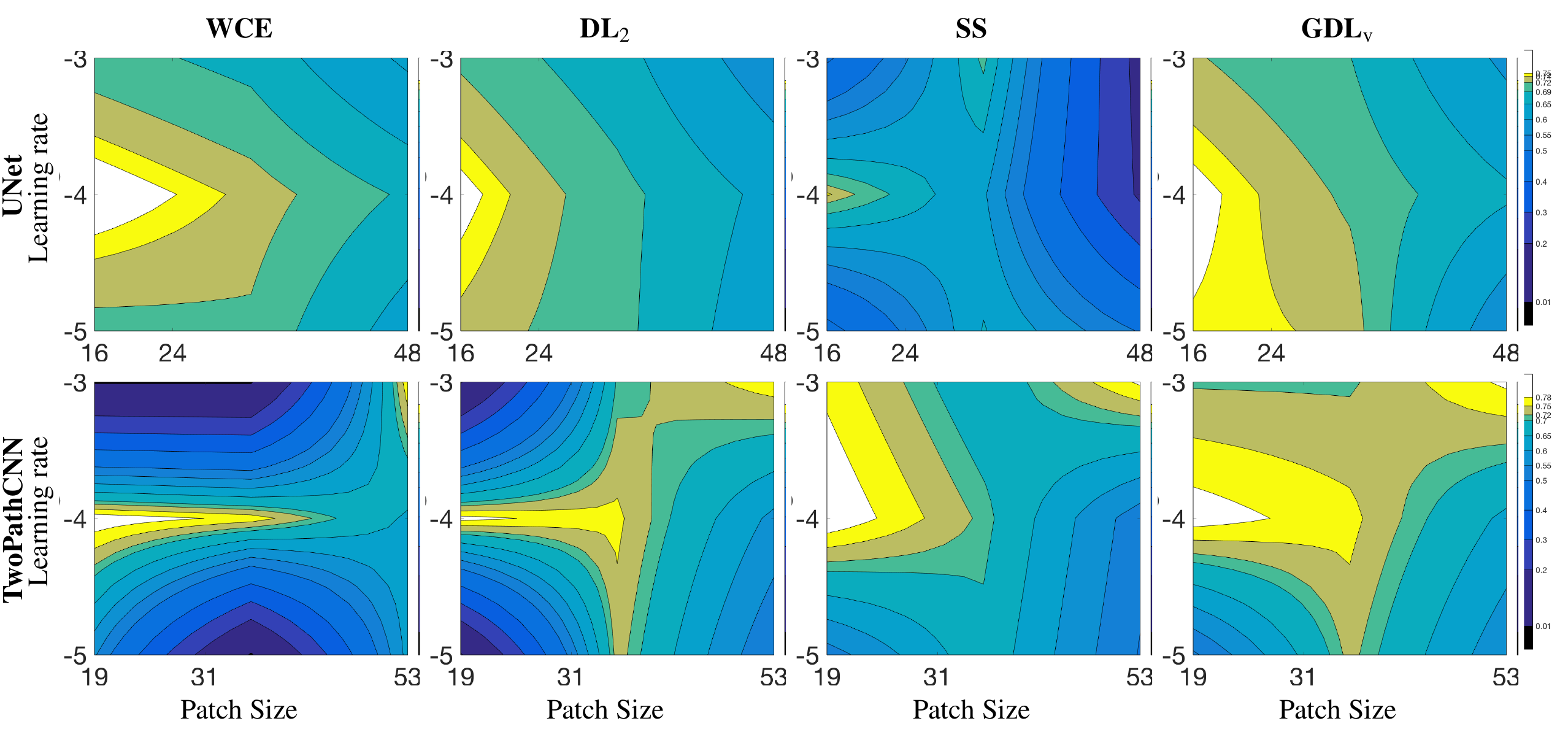}
   	\caption{Loss function behavior in terms of DSC (median over the last 200 iterations) under different conditions of effective patch size and learning rate in a 2D context. Isolines were linearly interpolated for visualization purposes.}
   	\label{fig:CompIsolines2D}
\end{figure}
 Table \ref{tab:2DResults} presents the statistics for the last 200 steps of training in term of DSC for the four loss functions at the different learning rates, and different networks while Figure \ref{fig:CompIsolines2D} shows the corresponding isolines in the space of learning rate and effective patch size illustrating notably the robustness of the GDL to the hyper-parameter space.
 The main observed difference across the different loss functions was the robustness to the learning rate, with the WCE and DL$_2$ being less able to cope with a fast learning rate ($10^{-3}$) when using TwoPathCNN while the efficiency of SS was more network dependent. An intermediate learning rate ($10^{-4}$) seemed to lead to the best training across all cases. Across sampling strategies, the pattern of performance was similar across loss functions, with a stronger performance when using a smaller patch but larger batch size.

 \subsection{3D Results}
   \begin{table}[pt]
   	\includegraphics[width=0.95\textwidth]{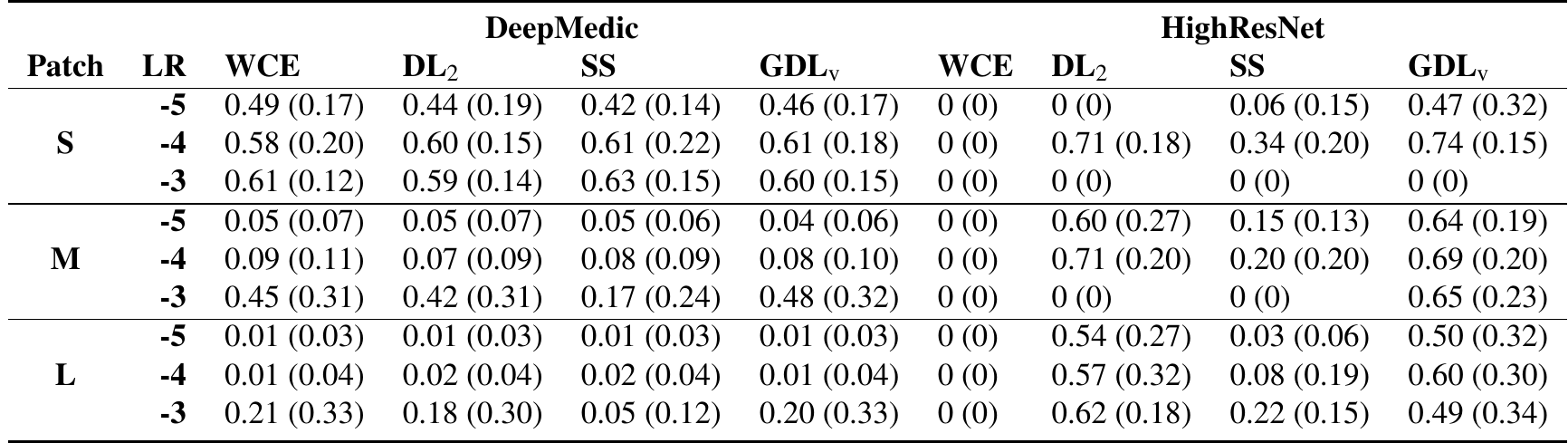}
   	\caption{Comparison of DSC over 200 last iterations in the 3D context for DeepMedic and HighResNet. Results are under the format median (interquartile range).}
   	\label{tab:3DResults}
   \end{table}

   \begin{figure}[pb!]
   	\includegraphics[width=0.9\textwidth]{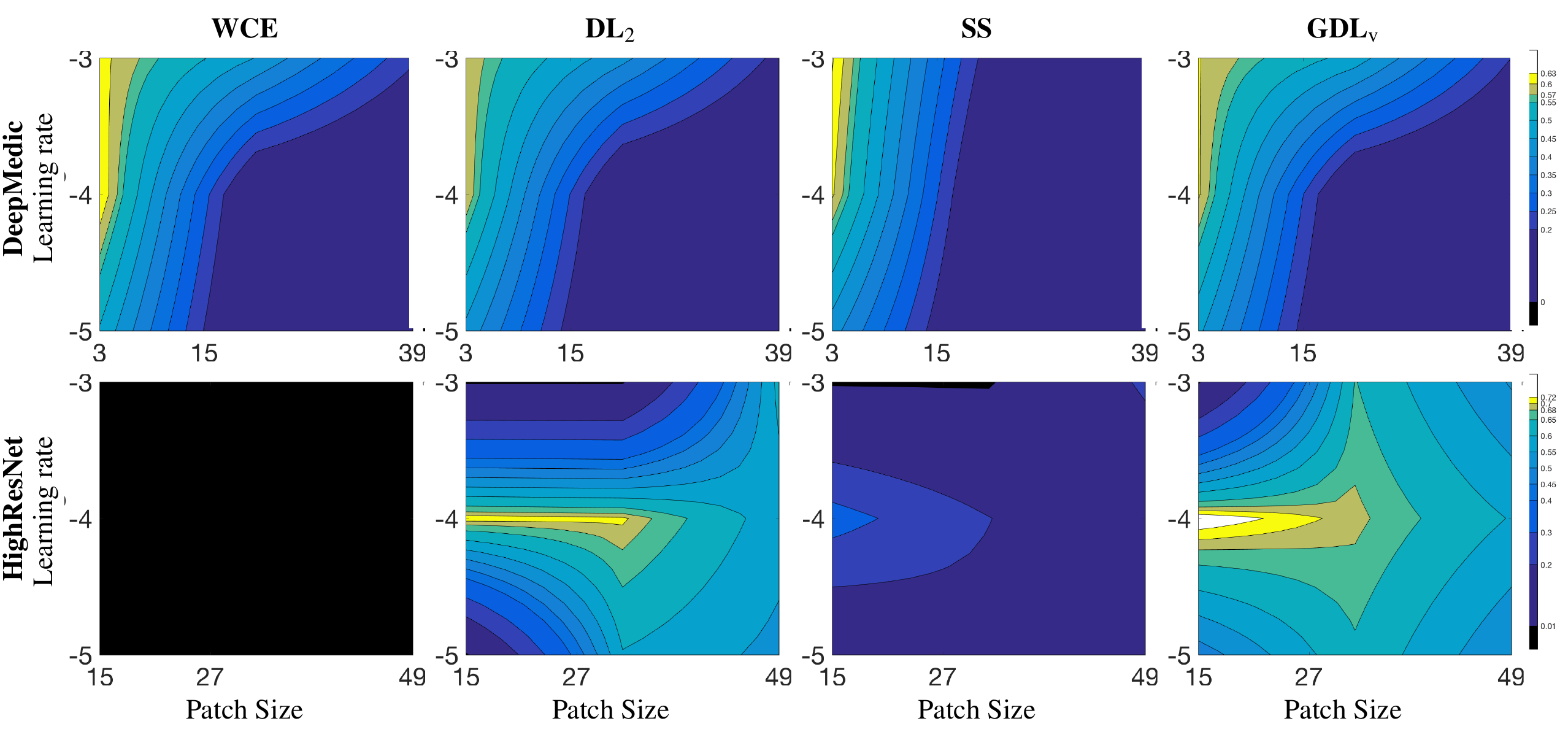}
   	\caption{Loss function behavior in terms of DSC (median over the last 200 iterations) under different conditions of effective patch size and learning rate in a 3D context. Isolines were linearly interpolated for visualization purposes.}
   	\label{fig:CompIsolines3D}
   \end{figure}
Similarly to the previous section,  Table \ref{tab:3DResults} presents the statistics across loss functions, sample size and learning rates for the last 200 iterations in the 3D experiment, while Figure \ref{fig:CompIsolines3D} plots the representation of robustness of loss function to the parameter space using isolines. Its strong dependence on the hyperparameters made DeepMedic agnostic to the choice of loss function.
In the 3D context with higher data imbalance, WCE was unable to train and SS dropped significantly in performance when compared to GDL$_v$. DL$_2$ performed similarly to GLD$_v$ for low learning rates, but failed to train for higher training rates. Similar patterns were observed across learning rates as for the 2D case, with the learning rate of $10^{-5}$ failing to provide a plateau in the loss function after 3000 iterations. We also observed that learning rates impacted network performance more for smaller patch sizes, but in adequate conditions (LR=$10^{-4}$), smaller patches (and larger batch size) resulted in higher overall performance.

\begin{figure}[pt]
    \includegraphics[trim=0cm 1cm 0cm 0cm, clip=true, width=0.35\textwidth]{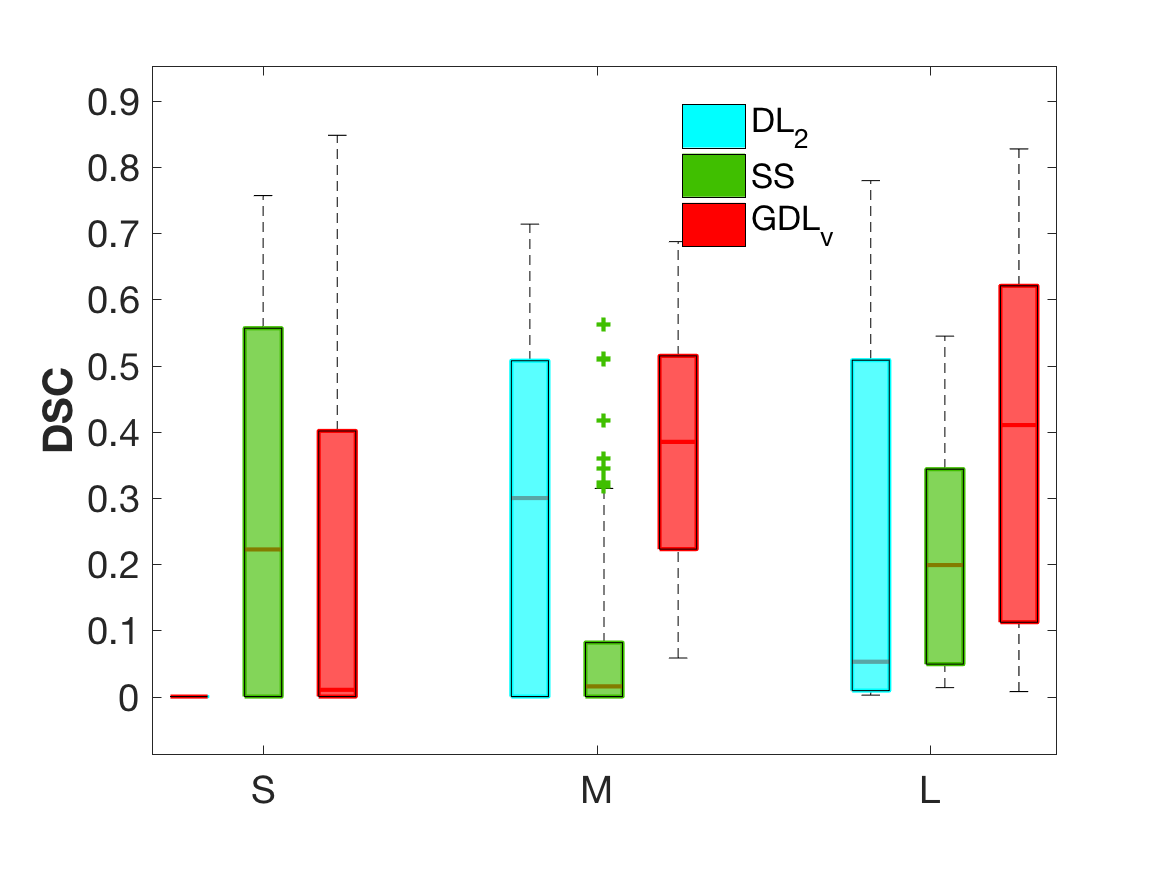}
     \includegraphics[trim=0cm 1cm 0cm 0cm, clip=true,width=0.35\textwidth]{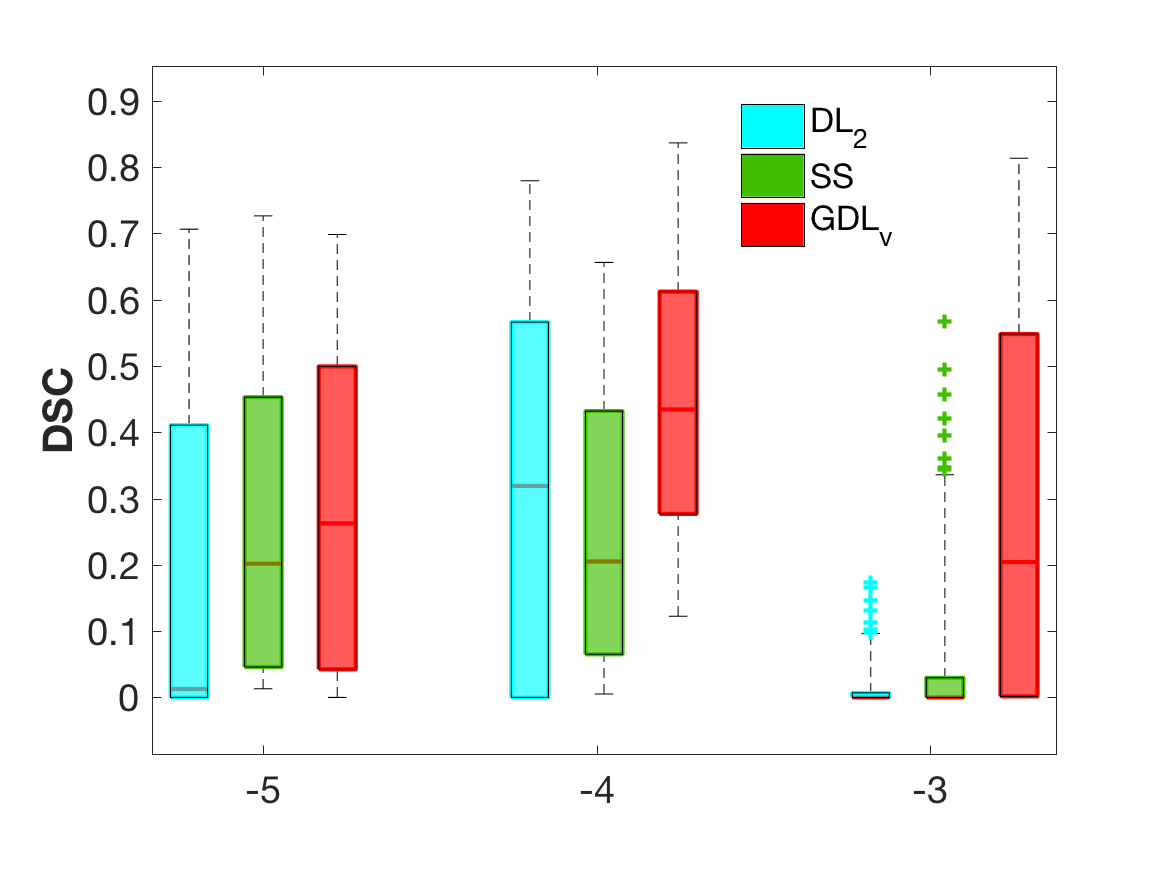}
    \caption{Test set DSC for all loss functions across patch sizes (left) and across learning rates (right). WCE was omitted as it was unable to cope with the imbalance.}
    \label{fig:DSCTest}
    \end{figure}

\begin{figure}[pb]
    \begin{tabular}{ccccc}
    FLAIR &Gold Standard &DL$_2$ & SS& GDL$_v$ \\
    \includegraphics[width=0.17\textwidth, trim=5cm 0cm 5cm 1cm,clip=true]{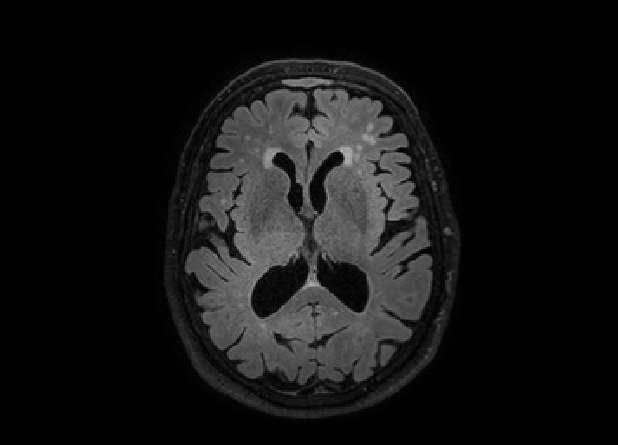} &
    \includegraphics[width=0.17\textwidth, trim=5cm 0cm 5cm 1cm,clip=true]{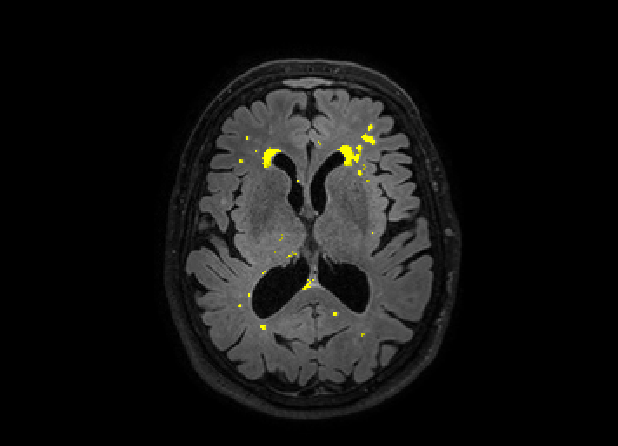}
    & \includegraphics[width=0.17\textwidth, trim=5cm 0cm 5cm 1cm,clip=true]{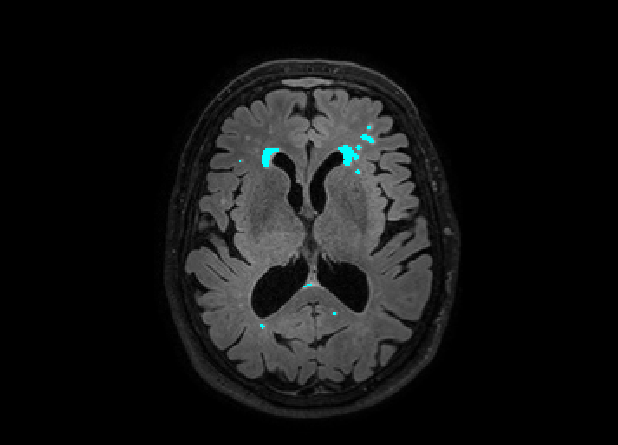}
&\includegraphics[width=0.17\textwidth, trim=5cm 0cm 5cm 1cm,clip=true]{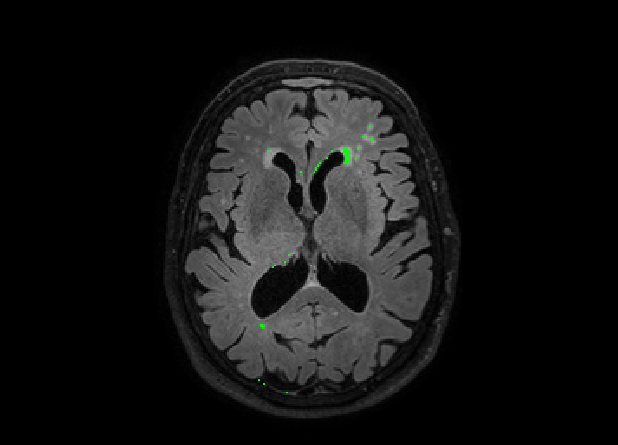}
&
\includegraphics[width=0.17\textwidth, trim=5cm 0cm 5cm 1cm,clip=true]{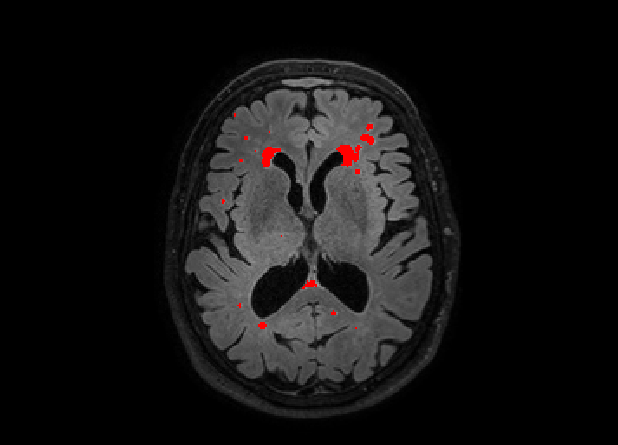}

    \end{tabular}
    \caption{The segmentation of a randomly selected 3D test set using different loss functions. Note the increased ability to capture punctuate lesions when using GDL$_v$. Loss functions were trained using a single patch of size $85^3$ per step at learning rate $10^{-4}$. }
    \label{fig:TestingResults}
    \end{figure}

\vskip -25pt\paragraph{3D test set}
For the 3D experiment, 10\% of the available data was held out for testing purposes. The final HighResNet model was used to infer the test data segmentation.  Figure~\ref{fig:DSCTest} shows the comparison in DSC across loss functions for the different sampling strategies (right) and across learning rates (left). Overall, GDL$_v$ was found to be more robust than the other loss functions across experiments, with small variations in relative performance for less unbalanced samples. 
    Figure~\ref{fig:TestingResults} presents an example of the segmentation obtained in the 3D experiment with HighResNet when using the largest patch size at a learning rate of $10^{-4}$.
\section{Discussion}
From the observation of the training behavior of four loss functions across learning rates and sampling strategies in two different tasks/networks, it appears that a mild imbalance is well handled by most of the loss strategies designed for unbalanced datasets. However, when the level of imbalance increases, loss functions based on overlap measures appeared more robust. The strongest reliability across setups was observed when using GDL$_v$. Overall this work demonstrates how crucial the choice of loss function can be in a deep learning framework, especially when dealing with highly unbalanced problems. The  foreground-background ratio in the most unbalanced case in this study was of 0.02\% for the 3D experiment (white matter lesions). Future work will focus on more extreme imbalance situations, such as those observed in the case of the detection of lacunes and perivascular spaces (1/100000), where deep learning frameworks must find a balance between learning the intrinsic anatomical variability of all the classes and the tolerable level of class imbalance. The studied loss functions are implemented as part of the open source NiftyNet package (\url{http://www.niftynet.io}).
\paragraph{Acknowledgments}
This work made use of Emerald, a GPU accelerated HPC, made available by the Science \& Engineering South Consortium operated in partnership with the STFC Rutherford-Appleton Laboratory. This work was funded by the EPSRC (EP/H046410/1, EP/J020990/1, EP/K005278, EP/H046410/1), the MRC (MR/J01107X/1), the EU-FP7 project VPH-DARE@ IT (FP7-ICT-2011-9-601055), the Wellcome Trust (WT101957), the NIHR Biomedical Research Unit (Dementia) at UCL and the NIHR University College London Hospitals BRC (NIHR BRC UCLH/UCL High Impact Initiative- BW.mn.BRC10269).
\label{sec:discussion}
\bibliographystyle{splncs03}
\bibliography{Bibliography338.bib}
\end{document}